\documentclass{fcs}
\pdfoutput=1
\usepackage{bm}
\usepackage{longtable}
\usepackage{indentfirst}
\usepackage{lipsum}
\usepackage{graphicx}
\usepackage{multicol}

\volumn{ }
\doi{ }
\articletype{REVIEW~ARTICLE}
\copynote{{\copyright} Higher Education Press and Springer-Verlag Berlin Heidelberg 2012}
\ratime{Revised December 04, 2022} 
\email{$ nasim.abdullah@ieee.org $}
\title{$\bm{The~ Prominence~ of~ Artificial~ Intelligence~ in~ COVID-19}$}
\author{MD Abdullah Al Nasim \xff $^{1}$, Aditi Dhali $^{1}$, Faria Afrin  $^{1}$, Noshin Tasnim Zaman  $^{1}$, Nazmul karim  $^{2}$, Anika Tabassum Sejuty $^{1}$, Kazi A Kalpoma $^{3}$, Md Mahim Anjum Haque $^{1}$}
\address{{1 \quad Department of Research and Development,
Pioneer Alpha}\\
{2\quad Electrical and Computer Engineering,
University of Central Florida} \\
{3\quad Department of Computer Science and Engineering,
Ahsanullah University of Science and Technology}} 

\begin{document}
\maketitle
\setcounter{page}{1}
\setlength{\baselineskip}{14pt}

\begin{abstract}
In December 2019, a novel virus called COVID-19 had caused an enormous number of causalities to date. While the front-line doctors and medical researchers have made significant progress in controlling the spread of the highly contiguous virus, technology has also proved its significance in the battle. Moreover, This survey paper explores the proposed methodologies that can aid doctors and researchers in adopting early and inexpensive methods to diagnose the disease. Most developing countries have difficulties carrying out tests using the conventional manner, but a significant way can be adopted with Machine and Deep Learning.
On the other hand, the access to different types of medical images has motivated the researchers. As a result, a mammoth number of techniques are proposed. This paper first details the background knowledge of the conventional methods in the Artificial Intelligence domain. Following that, we gather the commonly used datasets and their use cases to date. In addition, we also show the percentage of researchers adopting Machine Learning over Deep Learning. Lastly, in the research challenges, we elaborate on the problems faced in COVID-19 research, and after analyzing it, future directives are suggested to build an advanced, more developed health industry.
\end{abstract}


\section{Introduction}
After the flu of 1918, the novel Coronavirus or COVID-19 became the fifth documented outbreak and was discovered first in Wuhan; China \cite{li2020covid}. Gradually, it spread in the whole world. The International Committee on Taxonomy of Viruses gave the official name of the Coronavirus as "Severe Acute Respiratory Syndrome Coronavirus 2" or SARS-CoV-2 \cite{liu2020covid}. 
The World Health Organization (WHO) announced the SARS-CoV-2 outbreak as a Public Health Emergency of International Concern on January 30, 2020, and a pandemic on March 11, 2020 
It got reports of a cumulative total of 179,036,457 confirmed cases globally until October 13, 2022, according to PAHO.

\begin{figure}[t]
  \centering
 \includegraphics[width=9cm]{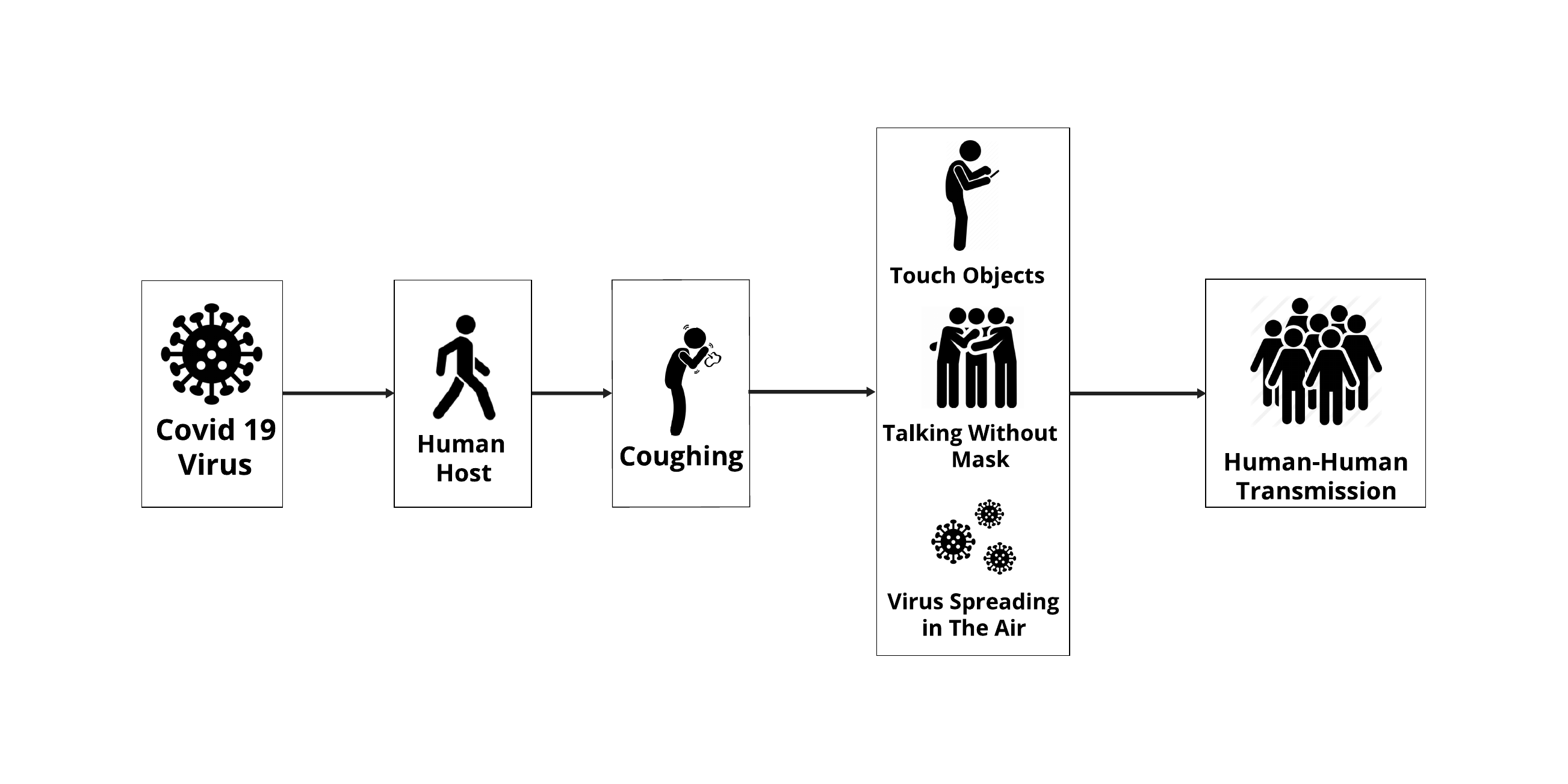}
  \caption{The Transmission Mechanism of COVID-19 from One Species to Another}
  \label{Covid transmission-01}
\end{figure}

If we look into the history, \cite{zhao2020potential}, considered bats as the natural host of SARS-CoV-2. Moreover, \cite{platto2021history} believed that bats are a locus for its expansion.
According to \cite{phan2020importation}, Coronavirus infection which arrived from China increased anxiety about human-to-human transmission. \cite{riou2020pattern} explained that human-to-human communication of 2019-nCoV is similar to the SARS-CoV of 2002. The authors suggested giving extra attention to the prevention of COVID-19. They showed that it is possible to end the epidemic according to the SARS-CoV of 2002 experience. In that case, the screening and control system at transportation stations and airports may play a vital role in preventing the outbreak of COVID-19. Fig. \ref{Covid transmission-01} shows the transmission of the Coronavirus.

From the ending of 2019, countries started to take precautions to prevent its outbreak. Though they started to maintain strict lockdown, curfew, inspire wearing face mask, regulate social distancing, the rate of positive COVID-19 cases and deaths increased significantly.

\begin{figure}[t]
  \centering
 \includegraphics[width=8cm]{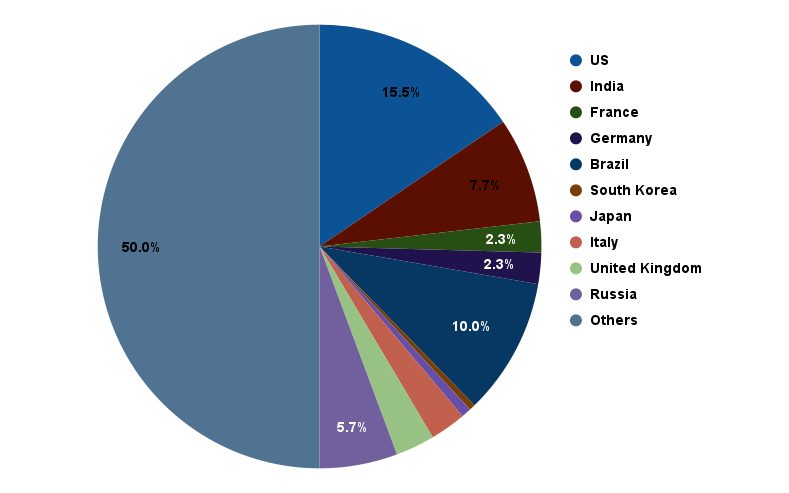}
  \caption{The COVID-19 case and death data of the Pie Chart was Collected from World Health Organization on December 4, 2022}
  \label{n1}
\end{figure}

Fig. \ref{n1} shows the percentage of Daily COVID-19 cases deaths on December 4, 2022, based on the top ten most affected countries so far. It is cleared that the number of total death case is more prominent in the United States than in the other nine countries now. The second most death cases are found in Brazil. After Brazil, there are the most founded death cases, respectively in South Korea and Japan but there death rate is comparatively pretty low than the other countries. The lowest percentage of death were in South Korea among these top ten countries which is 0.46\%. 

\begin{figure}[t]
  \centering
 \includegraphics[width=8cm]{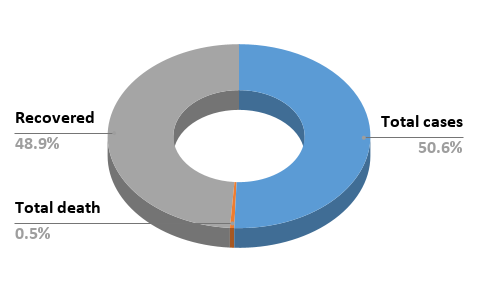}
  \caption{Total Affected Cases, Death Cases and Recovered Cases Around the World until 28 November, 2022}
  \label{deat-recover_chart}
\end{figure}

\begin{figure}[t]
  \centering
   \includegraphics[width=9cm]{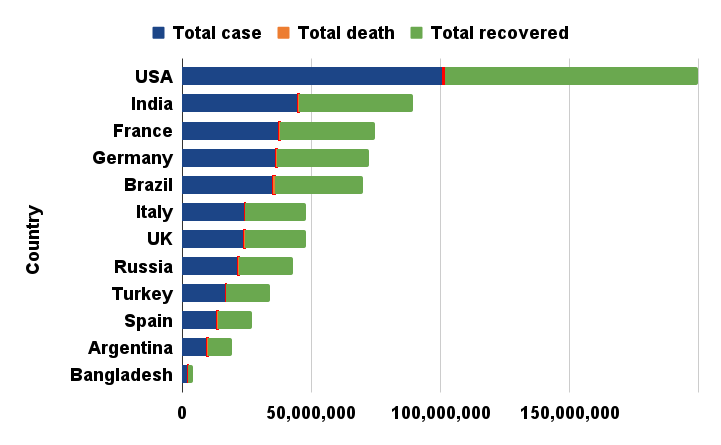}
   \caption{Total Affected Cases, Death Cases and Recovered Cases of Different Countries until 28 November, 2022}
   \label{case-death-recover_chart}
\end{figure}

Staring from the very beginning,within March 18, 2020, the number of COVID-19 patients and deaths were over 30,000 and 2500 \cite{spinelli2020covid}. At this point, the total number of affected cases, death cases, and recovered cases around the world from the very beginning of this pandemic till November 28, 2022, is depicted in Fig. \ref{deat-recover_chart}. Affected cases, death cases, and recovered cases based on different countries are shown in Fig. \ref{case-death-recover_chart}. We gathered these data from online (worldometer.info, New York Times, and pharmaceutical-technology.com).    
The United States of America tops this list with a total number of affected cases 97,618,392 when total death cases are 1,071,245 and the recovery rate is approximately 98.9\%.

\begin{figure}[t]
  \centering
 \includegraphics[width=8cm]{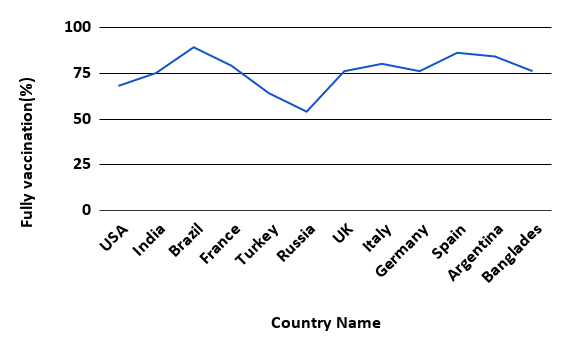}
  \caption{The Percentage of People Vaccinated with Respect to t heir Countries until 11 July, 2021}
  \label{vaccine_chart}
\end{figure}


Following the discovery of the COVID-19 pandemic, the importance of the models in overcoming COVID-19 challenges has multiplied. Taking these factors into consideration, in this paper, we have conceptualized and comprehensively represented a systematic review of the recent research on Machine Learning (ML) and Deep Learning (DL) in detecting COVID-19. This paper aims to effectively identify the ML and DL algorithms that helped in the early detection of COVID-19.

\begin{figure}[t]

 \includegraphics[width=9cm]{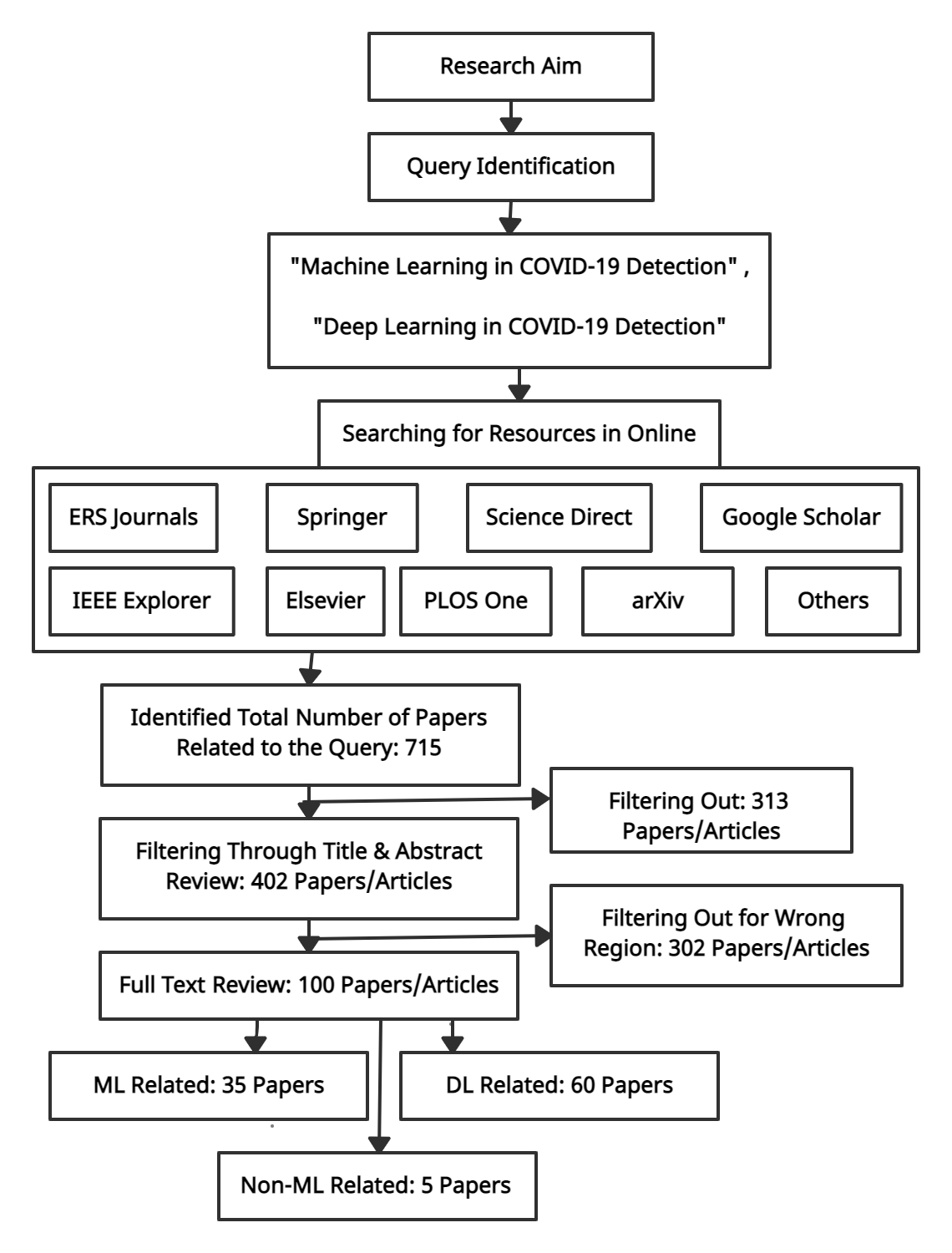}
 \centering
 \caption{Flowchart of Our Survey Work. It Demonstrates the Workflow from the Identification of Research Aim till the Collection of the Prominent Researches over the Recent Years in the COVID-19 Research Community}
 \label{flowchart}
\end{figure}

\begin{figure}[t]
  \centering
 \includegraphics[width=9cm]{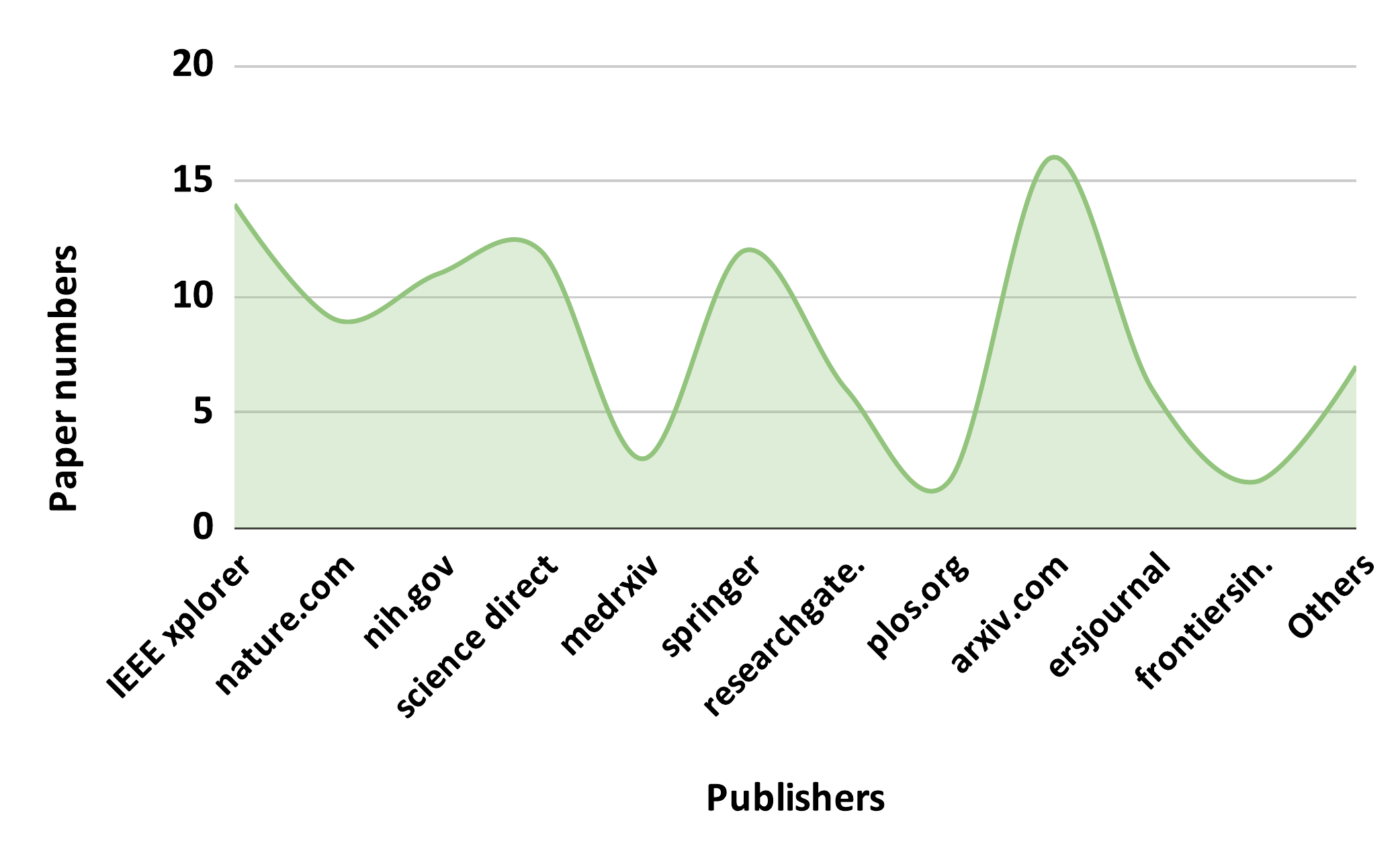}
  \caption{Plot to Demonstrate Our Collection of Research Papers from Different Renowned Publishers}
  \label{publisher_chart}
\end{figure}

The whole work we have done in this paper is depicted in this Fig. \ref{flowchart}. At first, we looked for research papers into famous online databases such as Elsevier, ERS journals, IEEE Xplore, Springer, Google scholar, Hindawi, MPDI, science direct, etc. On different websites, we have searched with some common keywords regarding the detection of COVID-19 using ML and DL methods and gather more than 700 papers. A graphical representation of different publishers and the number of documents we have used from those to develop this work has been depicted in Fig. \ref{publisher_chart}. We have filtered out more than 300 papers focusing on the documents that have been used for early detection or solving COVID-19 related issues. Exclusion procedures for filtering more than 250 publications that apply to COVID-19. Finally, we have reviewed a total of 100 papers related to ML and DL models in the detection of COVID-19 and some non-ML and DL algorithms that use image processing and computer vision algorithms in the early detection of COVID-19 and other COVID-19 related crises. In total, 100 papers, 35 papers used ML frameworks, 60 articles used DL algorithms, and five papers on non-ML and non-DL algorithms. We have separated them into different parts: methodology, summary, and the complete overview of other research work using tables. Data derived from each research paper evaluated for the analysis also include dataset used, the tool used, type of input used, augmentations technique, number of images used, accuracy, and other measures are highlighted.

\onecolumn
 
 


\begin{figure}[t]
  \centering
  \includegraphics[width=0.9\textwidth]{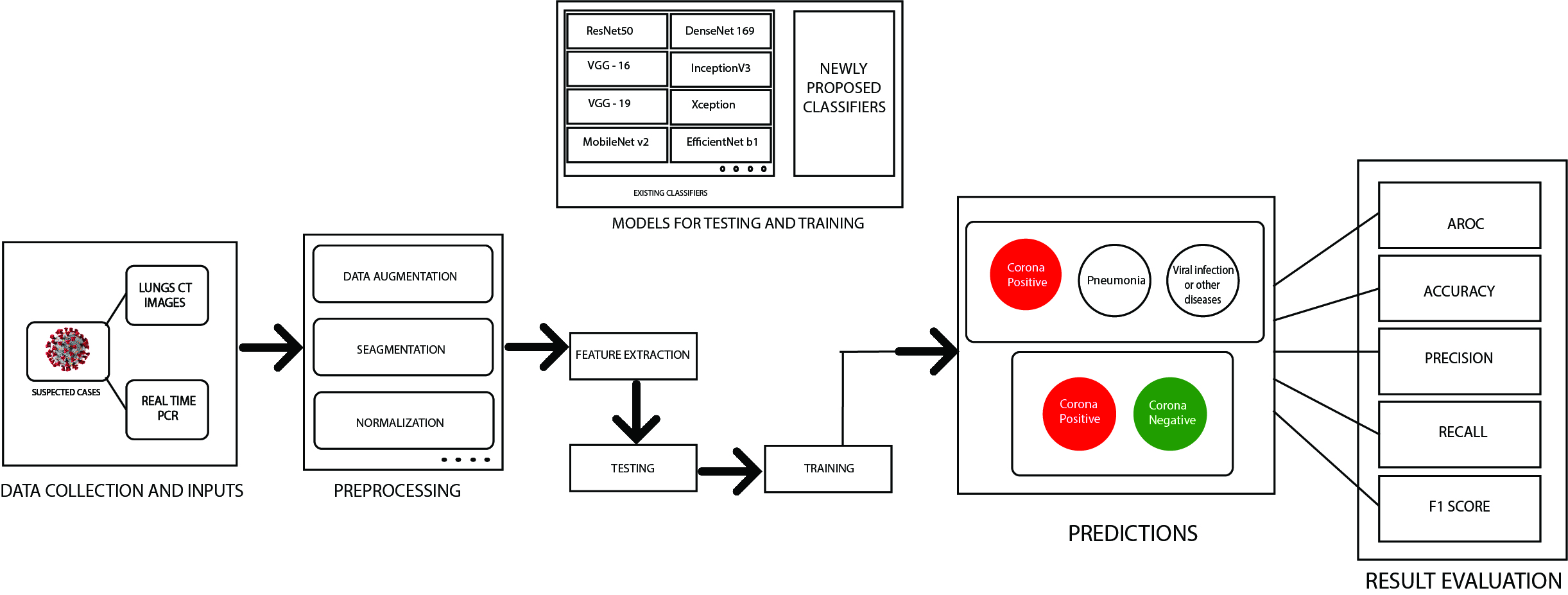}
 
 
  \caption{The Common Methodology of Deep Learning in Training and Evaluating Different Models with COVID-19 Datasets}
  \label{i7}
\end{figure}
\begin{multicols}{2}
\section{Background Study}
Deep learning is playing a big part in COVID 19 detection.
A significant number of strategies for identifying COVID 19 patients have already been developed by researchers. In Fig. \ref{i7}, we try to convey the core of the situation. We can state that we examine and follow this flowchart in most of the paper to draw their Coronavirus prediction utilizing DL.
Most of them use CT (Computed Tomography) or real-time reverse transcription-polymerase chain reaction (real-time RT-PCR) or CXR pictures of the lungs as input (Chest X-ray).
Though CT scans are not always as accurate as RT-PCR, they are less expensive, more accessible, more scalable, making them an intriguing choice for researchers. However, RT-PCR has a high specificity, its sensitivity changes with time. The data is then preprocessed using various data augmentation, normalization, and other procedures to improve the structure and produce a vast dataset. Many of the authors cited a tiny dataset as one of their constraints. These methods can assist one in dealing with specific situations. Following that, several features are retrieved to use the given strategy ideally. Data is separated into training and testing sets during the processing step. To carry out the tests, the authors utilized various ratios of training and testing datasets. They do the fine-tuning and cross-validation if the result is not satisfactory enough. Here, as the model, some used the existing DL classifiers, such as ResNet, EfficientNet-B0, MobileNetV2, DenseNet, Inception V3, or SqueezeNet, etc. Some other authors have applied their proposed model here. For example, Covid-resnet \cite{farooq2020covid}, Covid-caps \cite{afshar2020covid}, MetaCOVID \cite{shorfuzzaman2021metacovid}, Ensemble-CVDNet \cite{oksuz2020ensemble},  COVIDScreen \cite{singh2021covidscreen}, DL-CRC \cite{sakib2020dl}, Covid-fact \cite{heidarian2021covid} etc. As output, they reveal the prediction. Some predicted whether the input belongs to a COVID-19 patient or not. Some identified COVID 19 patients from other pneumonia patients or virus-affected patients. Evaluation or validation is done by accuracy, AUROC, F1 score, AUC, or recall.

AI approaches have the potential to provide significant benefits in combating COVID-19, but it is essential to consider the advantages against the potential risks and limitations.  AI algorithms can analyze vast amounts of data quickly and accurately, helping to identify patterns and trends that may be missed by humans. AI can predict the spread of COVID-19 and help authorities to plan and allocate resources efficiently.AI can help in drug discovery by analyzing vast amounts of data to identify potential treatments and vaccine candidates. AI requires vast amounts of data to function effectively, and there may be limited data available in the early stages of the pandemic, hindering AI's effectiveness. AI-based contact tracing and surveillance measures may raise privacy concerns among the public. AI-based contact tracing and surveillance measures may raise privacy concerns among the public. Careful consideration and collaboration between AI developers, healthcare professionals, and policymakers are essential to ensure that AI technologies are deployed effectively and ethically.


\section{Datasets}
Most scientific research depends on the collection and experiment of measurement data. Some datasets are private, and some are public for the authorities, politicians, researchers, and doctors. These scientific data sets are applied to develop new algorithms and differentiate those from other algorithms in the particular study. Nowadays, some datasets such as medical image data, electronic health records, country-wise data, public data, etc., are being used constantly for scientific research.

\subsection{Medical Image Data} Medical image data are gathered from X-rays, CT scans, MRI, intra-operative navigation, and biomedical research. It is the primary data source in medical science. 
Some renowned medical image data sources for COVID-19: Cohen Dataset, Praveen Corona Hack: Chest X-Ray Dataset, Mooney P. Chest X-ray Images (Pneumonia) Kaggle Repository, Pneumonia Dataset, CORD-19 dataset \cite{wang2020cord}.

\subsection{Electronic Health Record} Electronic health records, or EHRs, are real-time data collecting systems that save patient information in a digital format and make it accessible to authorized users quickly and securely. The patient's medical history, diagnosis, radiological images, and test results are all included. Medical errors are reduced, and patient data is protected, thanks to electronic health records. The most commonly utilized EHR datasets include Research Electronic Data Capture (REDCap), JSRT Dataset (Japanese Society of Radiological Technology), Covid-19 Radiography Database, and Italian Society of Medical and Interventional Radiology (SIRM) COVID-19 Database.

\subsection{Country Data} The country dataset is the resource where users get different variables for specific countries. 
Country data provides transparency and creates value to the community by making it publicly available. Some authors used country data in the papers:
\cite{zoabi2021machine} developed their model based on this. The Israeli Ministry of Health publicly disclosed dataset 
The dataset contains primary records of all the residents who were tested for COVID-19 countrywide. 
\cite{elsheikh2021deep} collected the Covid-19 recovered cases, confirmed cases, and deaths from the Saudi Ministry of Health website. They used the dataset as time-series data to train their proposed model.

USA based Covid-19 outbreak analysis was done by \cite{haratian2020dataset} and \cite{bashir2020correlation}. \cite{baqui2020ethnic}, \cite{de2021dataset}, \cite{cota2020monitoring}, \cite{wollenstein2020physiological}, \cite{silva2021using} analysed Covid-19 related data of Brazil. \cite{bhatnagar2020descriptive}, \cite{yadav2020data}, \cite{ulahannan2020citizen}, \cite{gupta2020trend}, \cite{bagal2020estimating} worked on Indian Covid-19 cases. Researchers studied bout country-wise characteristics, spreading rate, the mortality rate of Covid-19 in their research. 

Some other datasets are COVIDiSTRESS \cite{yamada2021covidistress} that analyze behavioral consequences of Covid-19, Geocov19 \ note {qazi2020geocov19} that consists of a vast number of multilingual Covid-19 related tweets. Melo et al. also came up with the first public dataset from Brazilian twitter \cite{de2020first}. This dataset extends the Covid-19 research from a different perspective.
Some popular datasets used in various research works related to COVID-19 are referenced in Table. \ref{Dataset table 1}. 

\section{Deep Learning in COVID-19 Battle}

Machine learning has begun to shine a light on hope in the face of the COVID-19 pandemic in this unclear and dire circumstance. ImageNet Classification, a CNN-based DL solution, earned significant success in the best-known international computer vision competition \cite{suzuki2017overview}, which sparked the adoption of ML in medical imaging area.
Experts utilise ML to anticipate the unknown future by predicting who is most likely to be affected, the danger of infection, diagnosing patients, developing treatments faster, and many other things. 

\subsection{Artificial Neural Networks In COVID-19}


\cite{wang2020deep} created a COVID-19 diagnostic DL model that was utilized on chest radiographs and categorized the X-ray images using transfer learning and combination-based COVID-Net. They developed ResNet-101 and ResNet-152 and trained on the ImageNet dataset according to the precision and loss rate after correcting and integrating the transfer learning model with a modified deep network layer.
\cite{rathee2021ann} has categorized the status of COVID-19 patients' health into infected, uninfected, exposed, and susceptible. They have used Bayesian and back propagation algorithms.

\cite{nath2020novel} used a Deep Neural Network to identify COVID-19 from x-ray pictures and CT scans using binary and multi-class classification in a study. Their proposed model is a 24-layer 6 CNN Network for X-ray and RT-PCR categorization. \cite{narin2020automatic} used a deep transfer learning-based strategy to predict COVID-19 patients using chest X-rays from four categories (normal, COVID-19, bacterial, and viral pneumonia). They used pre-trained transfer models based on deep convolution neural networks, which resulted in a high-accuracy decision assistance system for radiologists. A five-fold cross-validation procedure was applied to obtain a robust result. 80\% of the data was set aside for training, while the remaining 20\% was set aside for testing. With 96.1\%, ResNet50 and ResNet101 had the best overall performance. 

COVID-Net, a deep convolutional neural network design for detecting COVID-19 instances from chest X-ray (CXR) pictures, was introduced by \cite{wang2020covid}. COVIDx is an open-access benchmark dataset that contains 13,975 CXR images.
\cite{zhang2020covid19xraynet} employed a two-step transfer learning pipeline/strategy and the deep neural network architecture COVID19XrayNet to detect COVID-19 with only 189 annotated COVID-19 X-ray images. This COVID19XrayNet framework is created using a modified ResNet34 structure that outperforms the original ResNet34 design.


\cite{ghoshal2020estimating} attempted to determine how Bayesian Convolutional Neural Networks (Dropweights based) Networks (BCNN) can calculate uncertainty in DL resolutions to improve the diagnostic accuracy of the human-machine combination using COVID-19 chest X-ray in another paper. Their primary goal is to avoid false-negative detection. 

Author combined computational pharmacology, data mining, systems biology, and computational chemistry to find important targets, affected networks, and pharmaceutical intervention possibilities \cite{KANAPECKAITE2022106891}. The findings suggested pharmaceutical techniques consider therapeutic targets and networks. Data mining can reveal regulatory trends, identify novel targets, and flag undesirable impacts.

\subsection{Long Short-Term Memory Networks}
\cite{shi2021deep} proposed a DL-based quantitative CT model that incorporated five clinico-radiological factors. They considered it as more efficient tool than PSI and individual quantitative CT parameters for predicting the severity of COVID-19 patients. The selection operator LASSO logistic regression analysis is used to characterize the baseline of clinical-radiological features.
In one study, \cite{ abbasimehr2021prediction} proposed three hybrid methods for predicting the number of COVID-19 daily infected cases by combining different DL models, such as CNN-based method, multi-head attention-based method, and LSTM-based method with the Bayesian Optimization Algorithm. Using laboratory data, \cite{alakus2020comparison} developed a clinical predictive model to determine whether a patient could have Covid-19 illness. For model evaluation, they calculated precision, F1-score, recall, AUC, and accuracy ratings. Then, they put the model to the test using data from 600 patients in 18 labs. They next used 10-fold cross-validation and train-test split techniques to confirm it. They looked at samples from Israel's Albert Einstein Hospital. The authors trained six models: ANN, CNN, LSTM, RNN, CNNLSTM, CNNRNN, and two hybrid models: CNNLSTM and CNNRNN. 

\cite{devaraj2021forecasting} attempted to compare prediction models such as ARIMA, LSTM, Stacked LSTM, and Prophet techniques to forecast the spread of COVID-19 in terms of infected individuals, deaths, and recovered cases. They compared the forecasts' accuracy and picked the best model based on the various performance measurements and statistical hypothesis analysis.  \cite{jamshidi2020artificial} used DL models to forecast the number of Coronavirus positive reported cases for 32 states and union territories in India.
To forecast COVID-19 positive cases,

\end{multicols}
\begin{footnotesize}
\onecolumn
\begin{longtable}[h]{p{2.3cm}p{2.9cm}p{2.9cm}p{2.9cm}p{3.5cm}}
\caption{Commonly Used Dataset for Covid-19 Research}
\label{Dataset table 1}
\\ \hline \noalign{\smallskip}
  Serial No. & Name & Reference & Citations & Optimal Techniques\\
\noalign{\smallskip} \hline
\endfirsthead


\hline
\\
   1 & Cohen Dataset & \cite{cohen2020covid} & 491 & Local PatchBased Method, InstaCovNet19, VGG16, DenseNet-121 \\ \\
   
   2 & - & \cite{wang2017chestx} & 1374 & VGG16, ResNet-34, fficientNet-B0, ResNet50 \\  \\
   
   3 & JSRT Dataset & \cite{shiraishi2000development} & 620 & Local PatchBased Method \cite{oh2020deep}, Polynomial Regression \cite{punn2020covid} \\  \\
   
   4 & SCR Database & \cite{van2006segmentation} & 472  & Local PatchBased Method \cite{oh2020deep} \\ \\ 
   
   5 & MC Dataset & \cite{jaeger2014two} & 311 & Local PatchBased Method \\ \\ 
   
   6 & JHU CSSE Repository & - & - & KNN, SVM, DTs, Holt winter \cite{saba2021machine}, MLP \cite{car2020modeling}\\  \\
   
   7 & Covid-19 Radiography Database  & \cite{chowdhury2020can} & 197 & \cite{saba2021machine}, \cite{sahlol2020covid}, 24-Layer CNN \cite{nath2020novel}, \cite{waheed2020covidgan}\\  \\
   
   8 & - & \cite{rajpurkar2017chexnet} & 1064 & Inception with Fractional-order \cite{sahlol2020covid}\\  \\
   
   9 & Corona Hack: Chest X-Ray Dataset & - & - & Local PatchBased Method \cite{oh2020deep}  \\  \\
   
   10 & Pneumonia Dataset & - & 153 & DenseNet-121 \cite{rasheed2021machine}, \cite{sethy2020detection}, ResNet18 \cite{khalifa2020detection}\\  \\
   
   11 & COVIDx Dataset  & \cite{wang2020covid} & 661 & COVID-Net \\  \\
   
   12 & Dr. Adrian Rosebrock‚ Äôs Dataset  & \cite{hemdan2020covidx} & - & DenseNet201 \cite{hemdan2020covidx} \\  \\
   
   13 & RSNA Pneumonia Detection Dataset  & \cite{wang2020deep} & - & ResNet-101, ResNet-152, DenseNet201, Resnet50-V2, Inception-v3  \\  \\
   
   14 & Chest Imaging on Twitter & \cite{das2021automatic} & - & DenseNet201, Resnet50-V2, Inception-v3  \\  \\
   
   15 & COVID-19 Chest X-ray Dataset Initiative & \cite{das2021automatic} & - & DenseNet201, Resnet50-V2, Inception-v3  \\  \\

\end{longtable}
\end{footnotesize}
\twocolumn

 they used RNN-based LSTM variations such as Deep LSTM, Convolutional LSTM, and Bi-directorial LSTM.
The Convolutional LSTM model outperformed the other two models in the \cite{shastri2020time} study, with error rates ranging from 2.0 to 3.3 percent for all four datasets.
Table. \ref{Machine learning table} reflects some significant points of our work related to ML in COVID-19 patient detection. The parameters that Long Short-Term Memory Networks offer include learning rates, input and output biases, and a wide variety of others. It is substantially faster at training and generates a lot more successful runs. 

\subsection{Convolutional Neural Networks}
We categorize this subsection into three main parts. They are large scale networks, medium scale networks, and small scale networks. Each of them is again described in some smaller groups.

\subsubsection{Large Scale Network}

Due to the following epidemic, compiling a significant dataset to train a deep neural network has proven problematic. \cite{salman2020covid} sought to develop a DL model for detecting Coronavirus pneumonia on high-resolution X-rays, alleviating radiologists' workload and contributing to the epidemic's control. They defined the proposed solution as selected convolutional network architecture and examine the design decisions, evaluation methodologies, and implementation issues that go along with it. \cite{yoo2020deep} propose a deep learning-based decision-tree classifier for quickly recognizing COVID-19 from CXR pictures to make quick decisions about Covid patients. Three binary decision trees are used in their suggested classifier.
In a paper, \cite{hussain2021corodet} have proposed a novel CNN model which is based on 22-layer CNN architecture for the automatic detection of the virus as well as to find out the accurate diagnosis of two (COVID and Normal), three (COVID, Normal, and non-COVID pneumonia) and four (COVID, Normal, non-COVID viral pneumonia, and non-COVID bacterial pneumonia) class classification. 

The main objective of the paper of \cite{pham2021classification} was to find out the effectiveness of pre-trained CNN models for the detection of COVID-19 for rapid development. Thus, the classification tasks were fine-tuned using three public chest X-ray datasets and three pre-trained CNN models, namely SqueezeNet, AlexNet, and GoogleNet, without any data augmentation techniques. 



In \cite{jain2021deep}, After pre-processing three CNN models are employed to examine and compare their performance: ResNet, Inception V3, and Xception. 
Xception outperforms all three models. To improve existing COVID-19 detection using a limited number of publicly available CXR datasets, \cite{zebin2021covid} designed and developed a CXR-based COVID-19 disease detection and classification pipeline using a modified VGG-16, ResNet50, and new EfficientNetB0 architecture in this study.  



 \subsubsection{Medium Scale Network}
 
\cite{gupta2021instacovnet} proposed a DL-based Computer-aided diagnostic Framework (InstaCovNet-19) to diagnose COVID-19 accurately and with lower misclassification rates. This InstaCovNet-19 model is a DL algorithm built from various pre-trained models, including ResNet-101, Inception v3, MobileNet, and NASNet. 


 
Using an x-ray of the chest To detect the virus, \cite{car2020modeling} established a three-stage model. After preprocessing, they employed five state-of-the-art transfer learning models, AlexNet, MobileNetv2, ShuffleNet, SqueezeNet, and Xception, along with three optimizers, Adam, SGDM, and RMSProp, to reduce over-fitting. Finally, they have classified the data.




For the quick detection of the epidemic, \cite{panahi2021fcod} suggested a novel methodology called FCOD, which was based on the Inception Network. They employed 17 Depthwise Separable Convolution Layers in this study, and their automatic detection system can recognize complicated patterns from 2D X-ray images with a short forecast time.


\cite{sahlol2020covid} proposed a hybrid classification system dubbed Inception Fractional-order Marine Predators Algorithm, a blend of both Inception and Fractional-order Marine Predators Algorithm (FO-MPA) (IFM). 
This proposed method successfully extracted and selected features 136 for dataset one and 86 for dataset 2 out of 51K features with high accuracy.

\cite{heidari2021detecting}  developed a novel model incorporating preprocessing steps for the detection of COVID-19 pneumonia infected cases in their work and provided a quick decision-making tool for radiologists using computer-aided detection and diagnosis (CAD) systems.  
\begin{footnotesize}
\onecolumn

\begin{longtable}[h]{p{1cm}p{2.2cm}p{4.2cm}p{2.2cm}p{3.5cm}}
\caption{Details of Papers that Accommodated Shallow Machine Learning Techniques\label{Machine learning table}}
\label{table:Summary of the systems developed in recommender system, the techniques used}
\\ \hline\noalign{\smallskip}
References & Input Type & Method/ Technique Used & No. of Image Data & Augmentation Technique \\
\noalign{\smallskip} \hline
\endfirsthead

\\
\hline


\\
   \cite{callejon2021loss} & Text data & Comprehensive ML & 777 & NA \\ \\
    
   \cite {zoabi2021machine} & Text data & Gradient boosting machine & 51,831 individua-ls & NA \\ \\
    
   \cite{yao2020severity} & Clinical data & SVM, LR, RF, KNN, AdaBoost & 137 & NA \\ \\
    
   \cite{barstugan2020coronavirus} & CT & SVM & 150 & NA \\ \\
    
   \cite{chen2021machine} & CT & SVM & 326 & NA \\ \\
    
   \cite{zeng2021application} & Text data & NLP, XGBoost, MLP, Bi-LSTM & 301,363 & NA \\ \\
    
   \cite{rasheed2021machine} & X-ray & LR, PCA & 500 & Scaling \\ \\
    
   \cite{elaziz2020new} & X-ray & FrMEM, KNN & 2451 & NA \\ \\
    
   \cite{zhang2020viral} & X-ray & EfficientNet, ConfidNet & 519 & NA \\ \\
    
   \cite{cohen2020covid} & X-ray & NA & 679 & NA  \\ \\
    
   \cite{loey2021hybrid} & Image data & Resnet50 & 26140 & NA  \\ \\
    
   \cite{mele2021pollution} & Time series & TodaYamamoto test, ML  & Annual Data & NA  \\ \\
    
   \cite{kassani2020automatic} & Chest X-ray, CT & VGGNet, NASNet & 254 & NA  \\ \\
    
   \cite{sethy2020detection} & Xray & SVM, ResNet50, GLCM, AlexNet & 381 & NA  \\ \\
    
   \cite{khanday2020machine} & Text data & TF/IDF, BOW, LR & 212 & Preprocessing \\ \\
    
   \cite{sujath2020machine} & Time series data & LR, MLP & Not Defined & NA \\ \\
    
   \cite{patel2021machine} & Demographic and Clinical data & SVM, MLP & - & NA \\ \\
    
   \cite{tan2020study} & Clinical data, CT & Auto-ML & 319 & NA \\ \\
    
   \cite{saba2021machine} & Time series data & ARIMA, SARIMA, GBR, RF, KNN & - & NA \\ \\
    
   \cite{elhadad2020detecting} & Text data & DT, KNN, LR, LSVM, MNB, BNB & 7,486 & Text Parsing, Data Cleaning \\ \\
    
   \cite{wang2020weakly} & CT & AlexNet, UNet & 630 & Rando -maffine trans -formation, color jittering \\ \\
    
   \cite{car2020modeling} & Time series data & MLP  & 20706 Data Points & Conversion of data \\ \\
    
   \cite{yang2020covid} & CT image & U-Net & 812 & Gaussian blur \\ \\
    
   \cite{zoabi2021machine} & RTPCR result  & Grad-Boost & 99,232  & NA \\ \\
    
   \cite{pourhomayoun2021predicting} & Text & SVM &  2,670,000 & NA \\ \\
   
   \cite{heldt2021early} & EHR  & LR & 879 & NA \\ \\
   
   \cite{fernandes2021multipurpose} & EHR & ANN & 1040 & NA \\ \\
    
   \cite{shi2021deep} & CT image & LR & 196 & NA \\ \\
   \hline
   
   \end{longtable} \end{footnotesize} 
\begin{multicols}{2}




Supervised learning models need massive data to train and generalize. Swish-activated Residual U-Net GAN with dense blocks and skip connections. The recommended GAN architecture handles unpredictable brightness and generates realistic synthetic data due to instance normalization and sweep activation \cite{gulakala2022generative}. 

\textbf{Other Datasets and VGG:}
\cite{gianchandani2020rapid}  proposed an ensemble model for the identification of COVID-19 from infected, normal, and pneumonia people using two different transfer learning models, VGG16 and DenseNet. 
They tested the suggested model using two well-known datasets. For the early detection of COVID-19, \cite{qjidaa2020development} have proposed a clinical decision support system that consists of three phases based on the chest radiographic images. After the preprocessing and feature extraction, classification and prediction are made with a fully connected layer of several classifiers. Here, the VGG16 network is chosen to form this model. 

\textbf{Using CT Scan Datasets and GoogleNet:}
 
\cite{zhou2021ensemble} provide an article for identifying Covid-19 using 3 popular deep learning models named GoogleNet, ResNet, and AlexNet with transfer learning technique. They used preprocessed lung CT images to create 2500 high-quality pictures for training and validation.CNN's deep learning technology could revolutionize human life. \cite{KATHAMUTHU2023103317}This study uses deep transfer learning-based CNN to detect COVID-19 in chest CT images. Using VGG16, VGG19, Densenet121, InceptionV3, Xception, and Resnet50. This could help clinicians and academics build a therapy-decision tool.


\subsubsection{Small Scale Network}

\textbf{Using X-rays Image Datasets and DenseNet121:}

\cite{wang2020fully} presented a completely automatic innovative DL method to diagnose Covid-19 and perform a prognosis analysis using bare chest pictures without the use of human annotation in another publication. The planned DL system is divided into three sections. First, they employed a completely automatic Densenet121-FPN DL model for autonomous lung segmentation to avoid non-lung areas. Second, a non-lung area suppression operation was proposed inside the lung-ROI, with supplemental methods S4 employed to suppress non-lung area intensities.  Finally, they developed an auxiliary training procedure for pre-training to allow the COVID-19Net to learn lung features from a vast dataset using a DenseNet-Like structure.

\textbf{Using CT Scan Datasets and DenseNet:}

\cite{kamal2021evaluation} in a paper, tried to evaluate and made compare all the eight pre-trained models namely InceptionV3, NasNetMobile, ResNet50V2, DenseNet121, VGG19, ResNet50, MobileNet, MobileNetV2 for the efficient classification of chest X-ray images. DenseNet121 beats all other models to effectively classify Normal, COVID-19, SARS, Bacterial Pneumonia, and Viral Pneumonia based on chest X-ray pictures. \cite{panwar2020application} proposed a method called nCOVnet, a DL neural network-based fast screening (in 5 seconds) approach for identifying COVID-19 by analyzing X-rays, which can address two drawbacks of high cost and limited availability of testing kits. 

\textbf{Other Datasets and DenseNet:}

\cite{zhu2020deep} created a deep neural network architecture for feature selection that consisted of 6 fully linked dense layers and 56 clinical variables as inputs. The model prediction performance was utilized to rank the 56 features from the neural network by importance after fitting using permutation importance methods.

\textbf{Using CT Scan Datasets and AlexNet:}

\cite{turkoglu2021covidetectionet} built an expert system based on chest X-ray images dubbed the COVIDetectionNet model, achieving a 100\% success rate for the detection using pre-trained deep features ensemble as well as feature selection.  Firstly, a pre-trained AlexNet architecture is employed for feature extraction. Secondly, the most efficient features are chosen using the Relief Algorithm based on the K-Nearest Neighbor algorithm. Finally, these selected characteristics are classified using the Support Vector Machine Method. \cite{jain2020deep} developed a two-stage strategy model to detect the presence of COVID-19 while distinguishing COVID-19 induced pneumonia from healthy, bacterial pneumonia, and viral pneumonia cases in one study.  

\subsection{Graph Neural Networks }

\cite{khalifa2020detection}  recognized pneumonia chest x-ray using generative adversarial networks (GAN). The efficiency of the suggested model was reflected in GAN, with Resnet18 achieving the best accuracy. The proposed model is divided into three stages and is based on two DL models. 

\cite{toutiaee2021improving} compare multiple time-series prediction techniques incorporating auxiliary variables. Their Spatio-temporal graph neural networks approach forecast the course of the pandemic.

\cite{saha2021graphcovidnet} have proposed a Graph Isomorphic Network (GIN) based model named GraphCovidNet to detect Coronavirus by using CT-scans, and CXRs of the Covid affected patients.

By merging clinical, radiological, and biochemical data, \cite{lassau2021integrating} sought to predict hospitalized patients' outcomes. They used a DL model to assess CT scan images and a radiologist report that included a semi-quantitative description of CT scans to examine the increased amount of information offered by CT scans.


\cite{wang2020weakly} presented a weakly supervised 3D deep CNN dubbed DeCoVNet to Detect COVID-19 in a study for COVID-19 classification and lesion localization. Using the 2D UNet architecture, the first retrieved the 3D lung border of each chest CT volume for lung area segmentation. 
DeCoVNet was built using medical ground truth labels such as COVID-positive and COVID-negative. The DeCoVNet was trained to generate predictions using the testing data. 
They got the affected lesion localization result on the CAM activation region with the most overlap.

\cite{wang2021deep} attempted to develop an AI program that can effectively analyze representative CT images to screen Covid-19 using DL Method by modifying the typical Inception Network, based on the transfer learning neural network, and fine-tuning the modified Inception (M-Inception) Model with pre-trained weights.
They used a fully linked layer to perform classification and prediction.


\cite{zhang2020automated} used DL-based software (uAI Intelligent Assistant Analysis System) to aid in the diagnosis and quantification of COVID-19 Pneumonia. A modified 3D convolution neural network and a combined V-Net with bottleneck structures are used in this AI program. 

\



\subsection{Other DL Techniques}

\cite{wu2021deepgleam} presented DeepGLEAM. This coronavirus forecasting hybrid model beat existing ML models, such as the vector auto-regressive (VAR) model and a pure DL model, in one to four weeks forward prediction. 
They looked at approaches 
to quantify uncertainty. DeepGLEAM outperformed the competition in terms of accuracy and confidence intervals.
This article \cite{wan2021interpretation} developed a visual interpretative framework for understanding DL algorithms to diagnose the epidemic. Their approach generates a comprehensive interpretation of the DL model from various perspectives, including training patterns, diagnostic efficiency, trained features, feature extraction, hidden nodes, diagnostic prediction analysis, and so on.

\cite{pham2021deep} presented a framework for screening compound phenotype to determine the application for Covid-19 patients using a DL-based DeepCE model. 
Besides using a multiheaded attention mechanism they also suggest a new augmentation technique for extracting meaningful information from configurations that aren't useful. They underscore the model's importance by using it to re-purpose the drugs for the virus.

\cite{saba2021machine} developed a potential model for measuring and the optimal lockdown approach to alleviating the epidemic's causalities in one study. 
The over-fitting model authors studied three countries from each kind of type of lockdown.
\cite{zhang2020viral} utilized chest X-rays to screen for viral Pneumonia using CAAD, a machine-driven and accurate technique that relies less on labeled viral pneumonia data.
They wanted to improve the screening sensitivity by predicting anomaly detection failures.

\cite{tan2020study} proposed to develop and validate a Non-focus area prediction model in the early stages of Coronavirus Pneumonia. They used the Auto-ML method to excavate the texture features of the first chest CT image and assess the model's value in terms of the degree of Non-focus area damage and clinical classification.

\cite{saez2021potential} investigated significant biases from data source variability to uncover severity subgroups based on symptoms and comorbidities. They used the nCov2019 public dataset. 


\section{Critical Analysis on COVID-19 Research}

\subsection{Most common techniques acquired}

This chapter discusses the most commonly used framework of ML and DL Model techniques for COVID-19-related problems. An in-depth evaluation was performed, and the purposes for using these specific methodologies as well as their statistical data, are given in  Fig. \ref{Fig. 7}, Fig. \ref{Fig. 8} and Fig. \ref{Fig. 15}.Among all the papers we have reviewed, a significant percentage (60\%) of paper that has been used extensively is DL-based publications.


\begin{figure}[h!]
    \centering
   \includegraphics[width=8cm]{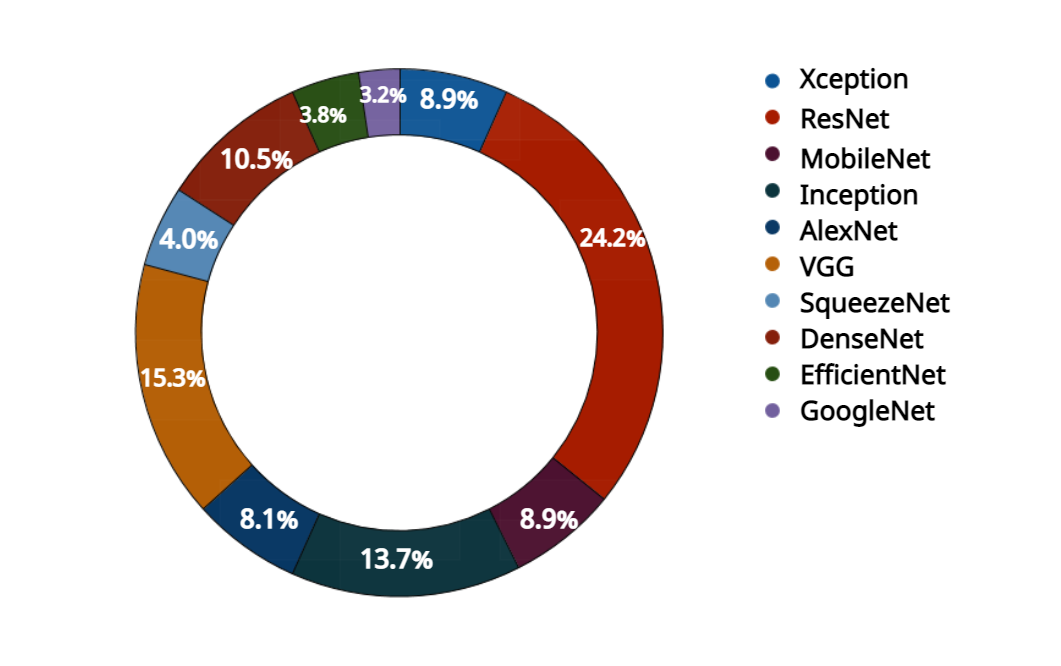}
   \caption{Chart demonstrating the usage percentage of CNN variants for COVID-19 research}
   \label{Fig. 9}
\end{figure}

We noticed that many ML papers were used for resolving COVID-19 issues out of all the documents included throughout this research study. We provided a survey of ML models' most used framework usage between them depending upon those ML publications. Looking at Fig. \ref{Fig. 7}, we can say that hybrid models (Deep LSTM, Convolutional LSTM, Bi-Directional LSTM, CNN-LSTM, CNN RNN, IRRCNN, BCNN), Decision Tree, SVM, RF and LR, XGBoost all these frameworks are primarily used in 8\% to 11.1\% of the comprehensive paper analysis. Other than utilizing these frameworks, SGD, NN, NLP, SVR, MLP, KNN, and other methods are used for about 7.4\% to 2.5\% papers. These frameworks are used extensively for classification and getting better and acceptable accuracy. Other ML methods, such as Ensemble Algorithm, Catboost, GNN, Gradient boosting, Grad-CAM, and SGD, were used in 1.2\% of the total paper. They have been utilized for statistical processing of COVID-19 data and also forecasting and predicting COVID-19 time-series data.


Correspondingly, for the framework that has been used in DL methods, we conducted an implementation review of the most commonly used techniques. A lot of frameworks, such as DNN, CNN, RNN, GAN, LSTM, and so on, are widely implemented, and the percentage of their usage is in Fig. 12. We found that CNN’s are the most commonly implemented DL methods for overcoming COVID-19 based challenges. CNN is the crucial framework that has higher accomplishment in the radiology, clinical imaging area. CNN is used widely in a lot of papers, almost 31\%. Moreover, we can see from Fig. \ref{Fig. 8} that LSTM was also primarily used in papers which are known as one of the most efficient approaches to DL algorithms for time series analysis of different data. 

Based on our study of  frameworks of CNN, one of the most diverse frameworks that have been used in most of the papers, we discovered that AlexNet, ResNet-34, ResNet50, MobileNet-V2, Inception V3, ResNet-18, U-Net, VGG-16, DenseNet, FC-DenseNet103, DenseNet121, GoogleNet, ImageNet, InceptionV3, Squeeze Net, and other models were used. From Fig. 9, ResNet algorithm is used mostly in maximum papers. VGG is the second most used framework, and DenseNet, AlexNet, XceptionNet CNN frameworks are most often employed for COVID-19 based disputes. 

\subsection{Commonly used Data}

With the new era of advancement, there is now tremendous change when we are getting the authentic source of data, and now the collection of the dataset is quite large. This continuous growth of medical images, it is quite difficult to analyze by a medical expert, may have come more variations from different expert and error-prone. 

For automating the diagnosis process, alternative solution is using ML and DL where these techniques have played an important role in medical image segmentation, image classification, and computer-aided diagnosis.
Based on this scenario, we have calculated the percentage of a total of 100 papers. After estimating percentage, it is noticed that image data is used in 63.6\% of papers, text data is used in 17\% of papers, time-series data is used in 10.2\% of papers, EHR data used in 5.7\% of paper, and other data forms are used in 3.4\% of papers. 
The results of this is shown in Fig. \ref{Fig. 15}. 

\subsection{ Machine Learning in COVID-19 Prediction and Detection}
According to the findings, computational ML, and DL algorithm techniques have been effectively used in the classification of clinical images through factors including the extraction of advanced features from spatial medical datasets and the capacity to differentiate between Sars-Cov-2 and viral pneumonia.
But the situation of pandemic’s ultimate uncertainty will mislead the goal of detection and prediction accuracy.

\end{multicols}


\begin{footnotesize}
\onecolumn
\begin{longtable}[h]{p{2.9cm}p{2.9cm}p{2.9cm}p{2.9cm}p{2.9cm}}
\caption{Details of the Papers that accommodate Deep Learning Techniques}
\label{tab:Deep_Learning_table}\\
\hline\noalign{\smallskip}
 References & Input Type & Method/ Technique Used & Number of Images Data & Augmentation Technique\\
\noalign{\smallskip} \hline
\endfirsthead

\caption{Continued From previous page}\\
\hline\noalign{\smallskip}
 References & Input Type & Method/ Technique Used & Number of Images Data & Augmentation Technique\\
\noalign{\smallskip} \hline
\endhead
\\
\hline
\endfoot

\hline

\endlastfoot
\\
   \cite{sethi2020deep} & Chest X-Ray & 7 Depth wise separable convolutions layers & 940 & Rotation, shifting, zooming, horizontal flipping  \\ \\
    
   \cite{nayak2021application} & Chest X-Ray & Xception, ResNet50, MobileNet, InceptionV3 & 6249  & NA \\ \\
    
   \cite{jain2021deep} & Chest X-Ray & InceptionV3, Xception, ResNet & 6432  & Rotation, Zooming, Horizontal flipping \\ \\
    
   \cite{oh2020deep} & Chest X-Ray & FC-DenseNet103, ResNet-18  & 502 & NA \\ \\
    
   \cite{jain2020deep} & Chest X-Ray & ResNet50, ResNet101  & 1215  & Rotation, Gaussian blur \\ \\
    
   \cite{wang2020fully} & CT Image & DenseNet121 -FPN, COVID-19Net  & 5372  & NA \\ \\
    
   \cite{bai2020artificial} & CT Image & EfficientNet B4 & 1186  &  lipping, scaling, rotation, brightness and blurring\\ \\
    
   \cite{punn2020covid} & Timeseries Data & SVR, PR, DNN, RNN, LSTM & 932,605 & NA \\ \\
    
   \cite{ahmed2021deep} & Chest X-Ray  & CNN & 1,389 & Rotation, shear, zoom, width shift, height shift \\ \\
    
   \cite{pham2021classification} & Chest X-Ray  & AlexNet, GoogleNet and SqueezeNet & 3666  & NA \\ \\
    
   \cite{xudeep} & CT Image & Modified Inception transferlearning model & 1,065 &  NA\\ \\
    
   \cite{hall2020finding} & Chest X-Ray & Transfer learming with VGG16 and Resnet50 & 455 & Horizontal flipping \\ \\
   
    \cite{abbasimehr2021prediction} & Time Series Data & $ ATT_BO $, $ CNN_BO $, $ LSTM_BO $ & Not defined & NA\\ \\ 
    
   \cite{panwar2020deep} & Chest X-ray, CTScan & Transfer learning with VGG-19, Grad-CAM & 5856  & NA \\ \\
    
   \cite{gupta2021instacovnet} & Chest X-Ray  & ResNet101, Xception, NASNet, InceptionV3, MobileNetv2 with an Integrated stack technique & 1088 & Shearing, zooming \\ \\
    
   \cite{qjidaa2020development} & Chest Radio -graphic Image & VGG16 & 300 & Random flipping, scaling, shearing \\ \\
    
   \cite{sahlol2020covid} & Chest X-Ray & FO-MPA &  3435  & NA \\ \\
    
   \cite{hussain2021corodet} & CT Scan, Chest X-ray & 22-layer CNN architecture & 7390  &  NA\\ \\
    
   \cite{gianchandani2020rapid} & Chest X-ray & Ensemble model using VGG16 and Dense-Net & - & Horizontal flipping, vertical flipping, rotation and sheer transformation  \\ \\
    
   \cite{kamal2021evaluation} & Chest X-ray & VGG19, InceptionV3, ResNet50, ResNet50V2, MobileNet, MobileNetV2, DenseNet121, NasNetMobile & 760 &  NA\\ \\
    
   \cite{zhang2020covid19xraynet} & Chest X-ray & Transfer learning, ResNet34 & 189  & Random vertical flip, resize crop, random horizontal flip and rotation \\ \\
    
   \cite{turkoglu2021covidetectionet} & Chest X-ray  & AlexNet, KNN, SVM & 6092 & NA \\ \\
    
   \cite{bai2020predicting} & CT Image & LSTM, MLP & 133 & NA \\ \\
    
   \cite{heidari2021detecting} & CXR Image  & VGG16  & 8,474 & Shearing, zoom, rotation, width and height shift, and horizontal flip \\ \\
    
   \cite{alakus2020comparison} & Text data & ANN, CNN, LSTM, RNN, CNN -LSTM, CNN -RNN & 600 & NA \\ \\
    
   \cite{jamshidi2020artificial} & RTPCR & Deep LSTM, Convolu -tional LSTM, Bi-direc -tional LSTM  & NA & NA \\ \\
    
   \cite{ghoshal2020estimating} & Chest X-ray & BCNN, ResNet50V2 & 6009 & Rotation, ZCA whitened, scaling, shifting, shearing, flipped horizontally and vertically \\ \\
    
   \cite{shastri2020time} & Time series data & Stacked LSTM, Bi -directioral LSTM, Convolu -tional LSTM, MAPE & NA &  NA\\ \\
    
   \cite{salman2020covid} & X-ray images & X-ray images & 260 &  NA\\ \\
    
   \cite{zhu2020deep} & Text data & Neural network predictive model & 220 & NA \\ \\
    
   \cite{elsheikh2021deep} & Time series data & LSTM, compared with ARIMA, NARANN  & 9048 & NA \\ \\
    
   \cite{karakanis2021lightweight} & Chest X-ray & cGAN, ResNet8, CNN & 1255  & misa -ligned and slightly rotated \\ \\
    
   \cite{fan2020inf} & CT images & Inf-Net, SemiInf-Net, RA, FCN8s, U-Net & 1700 & Resizing \\ \\
    
   \cite{wang2020deep} & Chest X-ray & ResNet-50, ResNet-101, ResNet-152 & 1853 & NA \\ \\
    
   \cite{abdani2020lightweight} & Chest X-ray  & MobileNet V1, MobileNet V2, MobileNet V3, ShuffleNet-V1, SquezeNet, Dark COVID-Net, SPP-COVID-Net & 2686 & NA \\ \\
    
   \cite{fan2020exploiting} & Chest X-ray & AlexNet, MobileNetv2, ShuffleNet, SqueezeNet, Xception & 148 & NA \\ \\
    
   \cite{sun2020adaptive} & CT images & VB-Net, GGO, XGboost, AFSDF, LR, SVM, RF, NN, ElasticNet & 2522  &  NA\\ \\
    
   \cite{shalbaf2021automated} & CT scans & Efficient Nets (B0-B5), NasNetLarge, NasNetMobile, InceptionV3, ResNet-50, SeResnet50, Xception, DenseNet121, ResNeXt50, Inception -ResNetV2 & 746  & Horizontally and vertically shift, random rotation, flip horizontally \\ \\
    
   \cite{rahman2020automated} & CCTV image  & Image processing and CNN & 858 & Resizing \\ \\
    
   \cite{nath2020novel} & Chest CT, X-ray images & CNN, GGO, Transfer learning & 4561 & Resizing, Max pooling \\ \\
    
   \cite{zeng2021application} & Text data & NLP, XGBoost, MLP, Bi-LSTM, SVR & 301,363  & NA \\ \\
    
   \cite{sandu2020application} & CT scans & FastCapsNet, CapsNet, ResNet50, AlexNet & 3000  & NA \\ \\
    
   \cite{das2021automatic} & X-Ray images & DenseNet201, ResNet50V2, Inceptionv3 & 1006  & Normalizing, resizing \\ \\
    
   \cite{jain2021deep} & X-Ray images & XCeption, InceptionNet3, ResNet, CNN & 6432 & Scaling \\ \\
    
   \cite{perumal2021detection} & CT scan, CXR images & Resnet50, VGG16, InceptionV3 & 5,232 & Max pooling 2d \\ \\
    
   \cite{fan2020exploiting} & X-ray images & AlexNet, MobileNetv2, ShuffleNet, SqueezeNet, Xception & 148 & NA \\ \\
    
   \cite{narin2020automatic} & X-ray images & InceptionV3, ResNet50, ResNet101, ResNet152, InceptionResNetV2  & 7065  & NA \\ \\
    
   \cite{punia2020computer} & images & ResNet-34, ResNet-50  & 1122 &  Rotating, zooming, height shifting\\ \\
    
   \cite{gozes2020rapid} & CT Scan Images & Resnet-50, Deep CNN  & 6356 & NA \\ \\
    
   \cite{brunese2020explainable} & Chest X-rays & VGG-16, Grad-CAM & 6,523  & Average Pooling 2D, Flatten, Dense, Dropout, Dense \\ \\
    
   \cite{li2020covid} & CXR images & FC-classifier, MobileNetV2, DenseNet-121, SqueezeNet & 537 & NA \\ \\
    
   \cite{mason2020pathogenesis} & NA & Patho-biology of SARS-CoV-2 virus & NA &  NA\\ \\
    
   \cite{lokwani2020automated} & CT scans & Xception, U-Net & 275 & NA \\ \\
   
   \cite{zhang2020automated}   & RT-PCR results, CT scans & It is a software & 4675  &  NA\\ \\
       
   \cite{barbosa2020automated} & CXR images & (1) Sectra PACS (2) ITK -Snap & 86 & AD segmentation\\ \\
       
   \cite{ghoshal2020estimating}    & X-ray images & ResNet50V2 & 6009 & Whitened, rotated, flipped horizontally, vertically, scaled, shifted, sheared \\ \\
       
   \cite{khalifa2020detection}    & X-ray images & AlexNet, GoogleNet, SqueezNet, ResNet18 & 5863  & NA \\ \\
       
   \cite{wang2020covid} & CXR images & VGG-19, ResNet-50  & 13,975 & NA \\ \\
       
   \cite{hemdan2020covidx} & Texts, tweets  & GNN & 6460  & Separating comma, other punctuations \\ \\
       
   \cite{hamid2020fake} & X-ray images & VGG19, DenseNet, InceptionV3, ResNetV2, Xception, MobileNetV2, InceptionResNet-V2  & 50 & NA \\ \\
       
   \cite{tu2020exploration} & Texts, articles  & ScispaCy NER model, BIONLP13CG corpus & 60,000 papers & Semantic visualization techniques \\ \\
       
   \cite{he2020sample} & CT scans & VGG-16 , ResNet-18 , ResNet-50, DenseNet-121, DenseNet-169, EfficientNetb0, EfficientNet-b1, CRNet & 746 & NA \\ \\
       
   \cite{panwar2020application} & X-ray images & VGG16 Model  & 6388 & Rotation, flipped horizontally and vertically  \\ \\
       
   \cite{alom2020covidmtnet} & CT image, X-ray  & IRRCNN & 5216  & Class specific data augmentation \\ \\
       
   \cite{song2021deep} & Images  & ResNet-50, DenseNet, VGG16  & 1990  & Translational and rotational \\ \\
       
   \cite{waheed2020covidgan} & CXR images & VGG16  & 1124 & Synthetic augmentation  \\ \\
       
   \cite{yoo2020deep} & CXR images & ResNet18 & 1884 & Horizontal flip, rotations, shifts\\

\end{longtable}
\end{footnotesize}
\newpage
\twocolumn


\begin{figure}[h!]
  \centering
 \includegraphics[width=8cm]{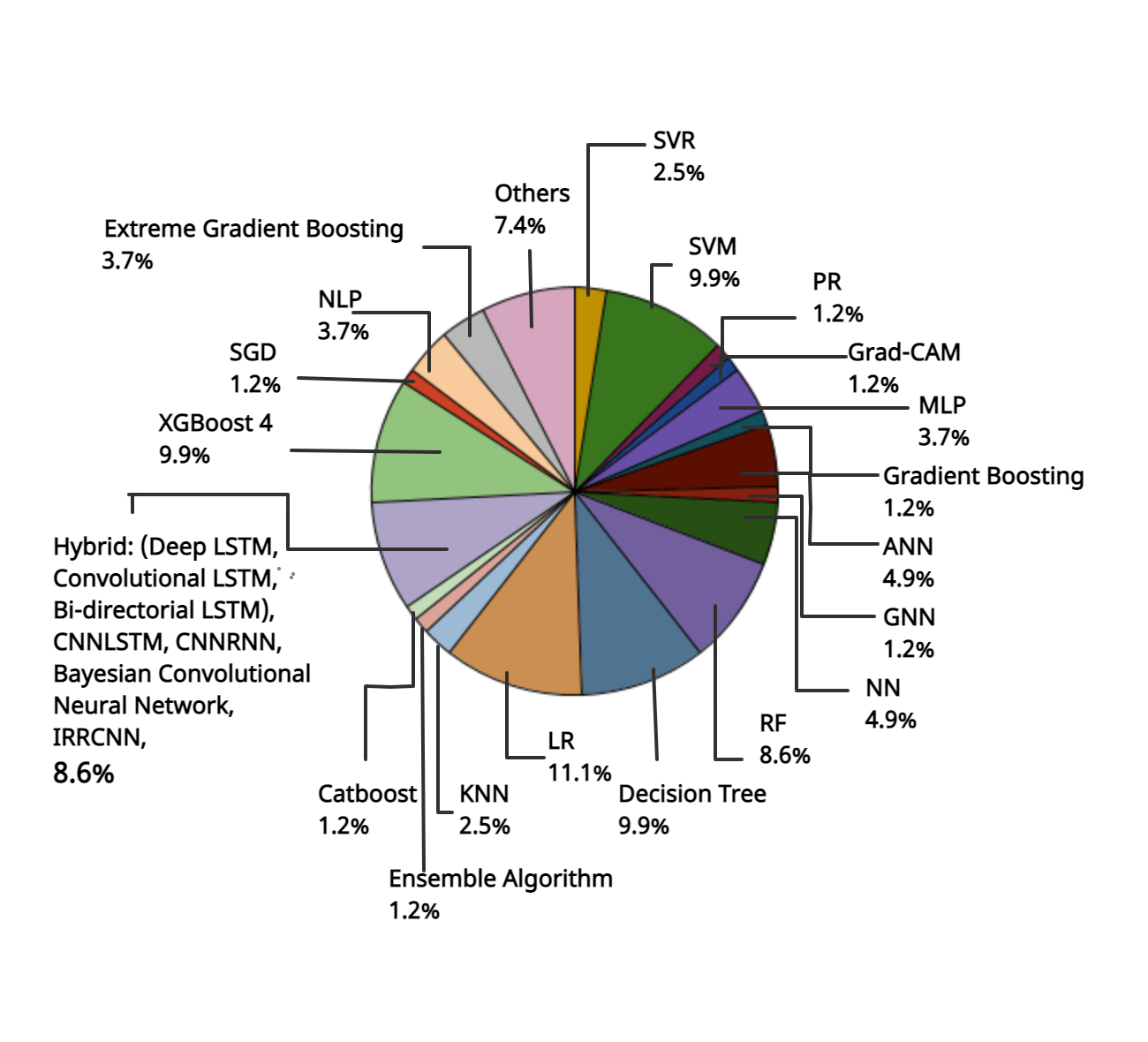}
  \caption{The percentage ratio of different types of Machine learning methods in COVID-19 research community}
  \label{Fig. 7}
\end{figure}


\begin{itemize}
    \item The exact amount of people affected died or still tested COVID-19 positive cannot be predicted with a 100\% accuracy because of the false-positive and false-negative high ratings and less COVID-19 detection test.
    
    \begin{figure}[h!]
  \centering
 \includegraphics[width=10cm]{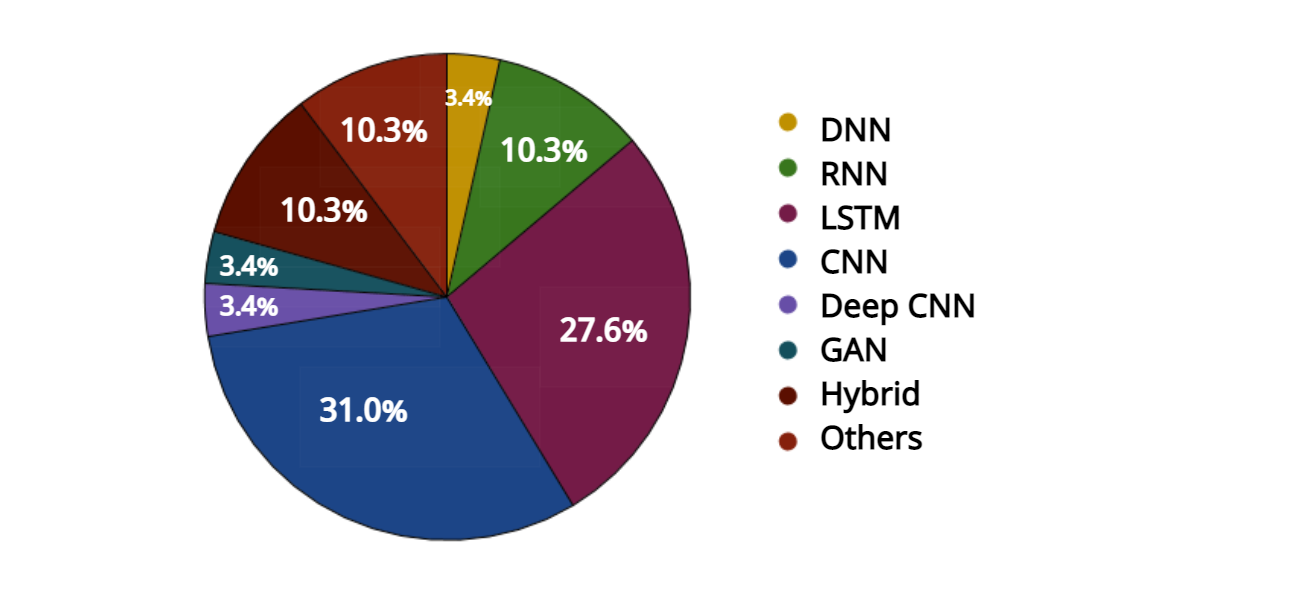} 
 \caption{The percentage ratio of different types of Deep learning methods in COVID-19 research community}
 \label{Fig. 8}
\end{figure}

\begin{figure}[h!]
  \centering
\includegraphics[height=120pt]{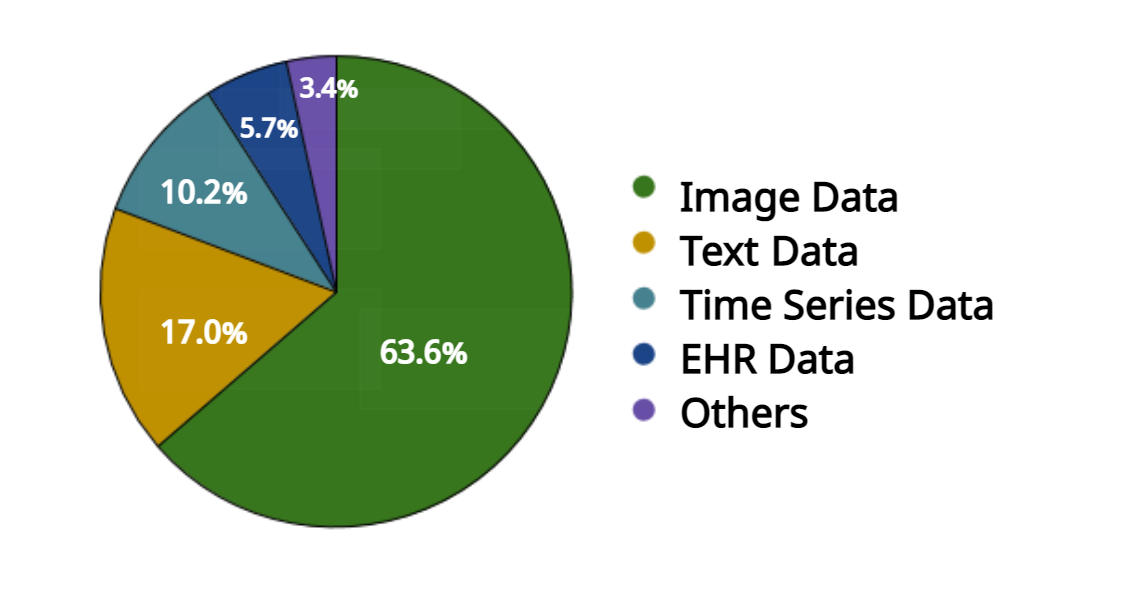}
  \caption{The percentage of published COVID-19 papers with respect to the types of data used in their research}
  \label{Fig. 15}
\end{figure}
    
    \item Proper annotation of images are also affecting the COVID-19 prediction using ML and DL. As most of the detection and prediction is done by medical images, annotation of the image dataset should be properly done. 
    \item The data that is currently available today is focused on COVID-19 tests that are conducted differently in various countries. They may not have to represent the whole viral infection in detail, which may be the ML and DL model's conceptual input. 
    To accurately diagnose and detect this disease, and to do proper forecasting using ML and DL methods we need to implement accurate data.\\

\end{itemize}
\subsection{Impact of Data Augmentation}

We can use data augmentation to get more variant data, prevent overfitting for better prediction, and perfectly train the model. 
By using data augmentation, it helps to increase the diversity of data as well as help to avoid overfitting. It can be more helpful for small datasets by increasing the size of data for better image segmentation, classification, and object detection. Before applying data augmentation, we need to understand the problem domain.

\begin{figure}[!h]
  \centering
   \includegraphics[height=120pt]{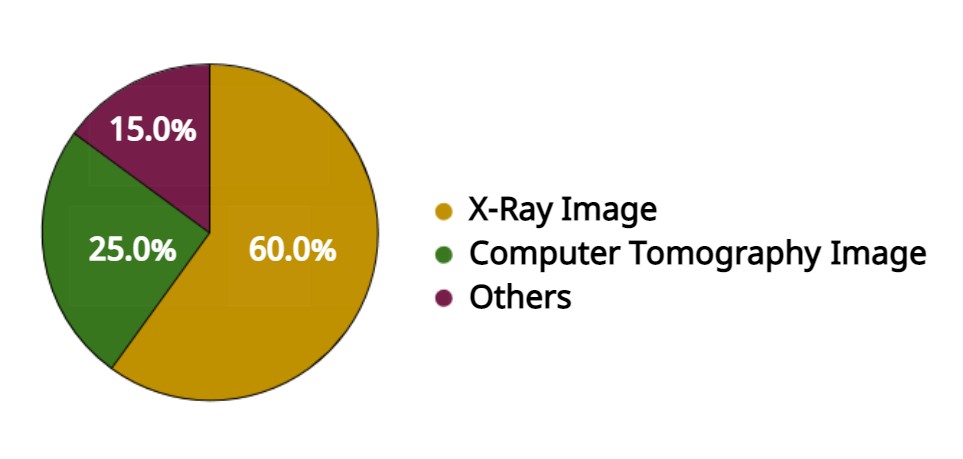}  
   \caption{Usage of types of images data that are acquired by our the paper demonstrated in our survey(Merge Fig.12 and 13)}
   \label{Fig.16}
\end{figure}

\begin{figure}[!h]
  \centering
 \includegraphics[height=150pt]{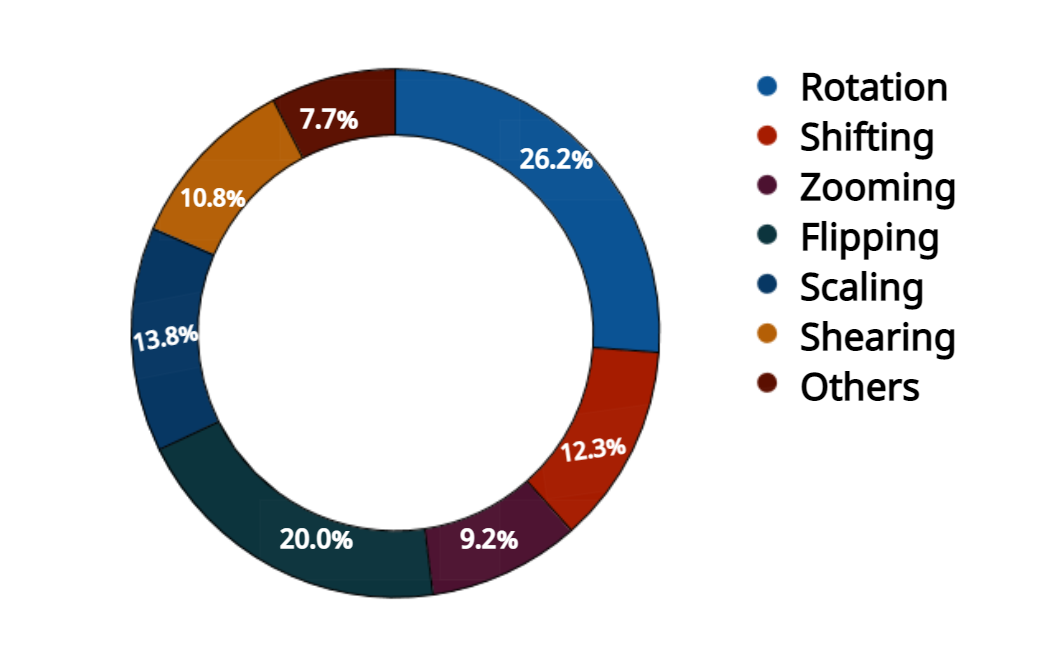} 
  \caption{Different types of Data augmentation techniques applied by the research papers illustrated in our survey}
  \label{Fig. 17DAG}
\end{figure}

It can be applied as a pre-processing step before train the model after estimating the percentage from Fig. \ref{Fig. 17DAG}, we can observe that powerful data augmentation techniques are used in many articles. Here, Rotation, Flipping, and Scaling have mainly used techniques. Also, there are having different data augmentation techniques along with others.
For developing a high-performance model, data augmentation can be used as a powerful tool as well as need rigorous problem analysis and domain knowledge.

\section{Research Challenges}

The following are the obstacles and issues related to DL implementations to monitor the COVID-19 pandemic:
\begin{itemize}
\item \textbf{Class-imbalance Problem:} As Covid-19 pandemic outbreaks in 2020, “data collection” was an obvious challenge for the authors. They find very little data on patients with the Coronavirus. On the other hand, vast data of pneumonia patients or patients with other lung diseases and healthy patients cause a class imbalance problem. To deal with this issue, \cite{zhang2020viral} suggested an anomaly detector instead of a classifier as the classifiers yield poor sensitivity performance for the class imbalance problem.

\item \textbf{Limited Data and Limitation Labeled Data:} Most of the papers published in 2020 faced this problem of limited data of COVID 19. They were trying to collect more data not only with quality but also with labeling and annotation.

\item \textbf{Lack of Data Quality:} Some of them collected data from online stores (books or pdf or newspapers) or from some random collection which impacts the quality of the research. Data should be collected from authentic sources like hospitals, clinics, or other research papers. Good quality data can make the analysis more structured. \cite{waheed2020covidgan} faced a challenge to create high-resolution samples from highly variable data sets. \cite{panwar2020application} have used only Posterior Anterior as a dataset.

\item \textbf{Selection Bias:} During collecting data, considering 4g different variables is a must to enhance the quality of the data. Such as \cite{cohen2020covid} did not consider immunosuppression status or ethnic background. For the primary research, it is a challenge to contemplate so many variables. Further research that is continuous analysis will improve this condition. Selection bias is also seen while gathering publicly available images.

\item \textbf{Less Number of Features:} Some of the papers demand that their work would be more efficient if the number of features were increased. They will cope up with this challenge in the further experiments they require.

\item \textbf{Single-center Design:} Single-center data impacts hugely in the testing. The result may vary in the large dataset or a multi-centered dataset. Due to the single-center design, their dataset is also smaller in size. \cite{barbosa2020automated} and \cite{zhang2020automated} found it difficult to collect multi-centered data. They felt that there was a need to do in-depth research to make this more potential.

\item \textbf{Dataset on X-ray, not CT image:} Comparing with computed tomography (CT), chest X-ray cannot come up with 3D anatomy and is less accurate than a PCR diagnostic. Chest Computed Tomography (CT) is a captivating option for patient management due to its spectacular portability, low cost,  availability, and scalability. It plays an essential role in identifying earlier lung infections and monitoring disease development. 
\item \textbf{Unbalanced Data:}  For improved training in DL models and to obtain better features, we need a vast amount of balanced dataset. Due to this problem, the authors encountered many hurdles and had to apply different strategies to make it balanced, which is time-consuming.
\item \textbf{Hardware Limitations:} Extensive data need for training intense models in less amount of time. Accordingly, we need more powerful GPUs and high computing power.

\item \textbf{Lack of Diverse Datasets:} Due to the lack of a diverse dataset, it is hard to analyze and perfectly learn the models. When the model is fed with various training data, it provides more discriminative information, making the model unique.
\item \textbf{Security of Privacy:} By preserving privacy and maintaining the security of patient’s medical information, it is hard to collect a large number of medical images, which hinders the training and validation of models.
\item \textbf{Time-intensive and Costly:} High-quality data preparation is needed for the fast detection of COVID-19 when there are hidden biases in the dataset. Typically it requires time-intensive, expensive human labor.

\end{itemize}

\section{Future Research Directions}

For fighting against the COVID-19, the new revolution of technology can be the savior of our lives. Future COVID-19 research directions will guide healthcare planning, telemedicine, management strategies, and robust disease prediction at the earliest stage. In this regard, AI-based ML and DL systems contribute to the available resources where Transfer Learning plays a significant role with the help of unbalanced and limited datasets. 
\begin{itemize}
\item \textbf{Robust Development Architecture for COVID-19 and other Diseases:} An automated, robust architecture can be built with the help of image classification using X-ray and CT images for the detection of COVID-19 as well as other diseases, which will provide an end-to-end solution for fast detection.
\item \textbf{Necessity of Combined Data Repository:} For further research development work, we need more data to combat the COVID-19 situation. Besides, this Combined Data Repository will help us to fight back with similar pandemics.
\item \textbf{Implementation of Medical Assistive in Real-World:} For supporting the health professionals, it is mandatory to implement this medical-assisting tool in hospitals, radiology clinics.
\item \textbf{Intelligent Robot:} For assisting the COVID-19 infected patients without human labor, these robots are supposed to be used in hospitals to curb the spread of disease.
\item \textbf{Coronavirus detection by checking breathing sound:} This can be a new and easier area for detecting coronavirus. Scientists can focus on this for  fast detection.
\item \textbf{AI in Biological Research:} For biological research, this AI system can be used to predict the structure of proteins and genetic sequence.
\end{itemize}
It can be stated that more future direction and research are required for efficient identification and diagnosis.



\section *{Appendixes}

{\noindent \em Remainder omitted in this sample. See http://www.jmlr.org/papers/ for full paper.}



\bibliography{source.bib}
\bibliographystyle{plain}

\end{document}